\documentclass[a4paper,conference]{IEEEtran}
\IEEEoverridecommandlockouts
\usepackage{cite}
\usepackage{amsmath,amssymb,amsfonts}
\usepackage{algorithmic}
\usepackage{graphicx}
\usepackage{textcomp}
\usepackage{xcolor}
\usepackage{comment}

\usepackage[caption=false,font=footnotesize]{subfig}  % add this in the preamble
\usepackage{tabularx} % add in preamble
\def\BibTeX{{\rm B\kern-.05em{\sc i\kern-.025em b}\kern-.08em
    T\kern-.1667em\lower.7ex\hbox{E}\kern-.125emX}}
\begin{document}

\title{Toward Accessible Dermatology: Skin Lesion Classification Using Deep Learning Models on Mobile-Acquired Images
}
%Automated Skin Lesion Classification Using Transformer-Based Deep Learning Models

\author{\IEEEauthorblockN{Asif Newaz}
\IEEEauthorblockA{\textit{Dept. of Electrical and Electronic Engineering} \\
\textit{Islamic University of Technology}\\
Gazipur, Bangladesh \\
eee.asifnewaz@iut-dhaka.edu}
\and
\IEEEauthorblockN{Masum Mushfiq Ishti}
\IEEEauthorblockA{\textit{Dept. of Electrical and Electronic Engineering} \\
\textit{Islamic University of Technology}\\
Gazipur, Bangladesh \\
masummushfiq@iut-dhaka.edu}
\and
\IEEEauthorblockN{A Z M Ashraful Azam}
\IEEEauthorblockA{\textit{Dept. of Electrical and Electronic Engineering} \\
\textit{Islamic University of Technology}\\
Gazipur, Bangladesh \\
ashrafulazam@iut-dhaka.edu}
\and
\IEEEauthorblockN{Asif Ur Rahman Adib}
\IEEEauthorblockA{\textit{Dept. of Electrical and Electronic Engineering} \\
\textit{Islamic University of Technology}\\
Gazipur, Bangladesh \\
asif-ur-rahman@iut-dhaka.edu}
}

\maketitle

\begin{abstract}

Skin diseases are among the most prevalent health concerns worldwide, yet conventional diagnostic methods are often costly, complex, and unavailable in low-resource settings. Automated classification using deep learning has emerged as a promising alternative, but existing studies are mostly limited to dermoscopic datasets and a narrow range of disease classes. In this work, we curate a large dataset of over 50 skin disease categories captured with mobile devices, making it more representative of real-world conditions. We evaluate multiple convolutional neural networks and Transformer-based architectures, demonstrating that Transformer models, particularly the Swin Transformer, achieve superior performance by effectively capturing global contextual features. To enhance interpretability, we incorporate Gradient-weighted Class Activation Mapping (Grad-CAM), which highlights clinically relevant regions and provides transparency in model predictions. Our results underscore the potential of Transformer-based approaches for mobile-acquired skin lesion classification, paving the way toward accessible AI-assisted dermatological screening and early diagnosis in resource-limited environments.

\end{abstract}


\section{Introduction}

Skin diseases are among the most common health problems worldwide, impacting millions of individuals across the world. According to the World Health Organization(WHO), more than 900 million people are affected by various skin conditions, with skin disorders ranking as the fourth leading cause of nonfatal disease burden worldwide \cite{b1}. These diseases range from mild infections to life-threatening chronic disorders. Despite their prevalence, timely and accurate diagnosis remains a major challenge, particularly in low-resource and rural settings where access to specialized dermatological care is limited. 

Skin diseases are typically identified through visual examination of the affected area by trained dermatologists. Clinicians rely on features such as the size, shape, color, texture, and borders of skin lesions to distinguish between different conditions. In many cases, dermoscopy, a non-invasive imaging technique, is utilized, which provides enhanced visualization of subsurface structures and improves diagnostic accuracy. For suspicious or ambiguous lesions, additional procedures such as biopsy and histopathological analysis are performed to confirm the diagnosis. However, these methods require specialized expertise and equipment, which may not be readily available in low-resource settings. Moreover, the process is often time-consuming and expensive. Dermatologists are not always available in all regions, particularly in rural areas, which further highlights the need for automated diagnostic solutions.

In recent years, deep learning (DL) models have emerged as powerful tools for the automated diagnosis of skin lesions from medical images \cite{b2}. Unlike traditional computer-aided diagnosis (CAD) systems that rely on handcrafted features, DL models can automatically learn hierarchical representations of visual patterns directly from raw images. Convolutional Neural Networks (CNNs) have been widely applied in dermatology, demonstrating strong performance in detecting and classifying skin lesions \cite{b3}. These models can capture discriminative features such as color variations, irregular borders, and textural differences that are critical for distinguishing benign from malignant conditions.

%Despite their success, CNNs often struggle to capture long-range dependencies and global contextual information, which are essential for accurately analyzing complex lesion structures.

One of the major limitations in applying DL methodologies is the requirement for large, well-annotated datasets. In medical applications, it is even more difficult to obtain a wide variety of data, as cases often lack diversity across demographics, imaging conditions, and rare disease types. However, the release of public skin lesion datasets has played a crucial role in advancing research on automated skin lesion identification. Among the most widely used is the International Skin Imaging Collaboration (ISIC) archive, which has organized annual challenges providing thousands of dermoscopic images annotated by experts \cite{b4}. Another important resource is the HAM10000 dataset, consisting of 10,015 dermatoscopic images across seven diagnostic categories, making it one of the most diverse benchmarks for skin lesion classification \cite{b5}. Smaller datasets such as PH2, Derm7pt, and MED-NODE have also contributed to the development and evaluation of CAD systems \cite{b6}.

%These publicly available datasets not only enable the training of robust deep learning models but also foster comparative evaluation and reproducibility across different studies.

One major limitation of most publicly available skin lesion datasets is that they primarily consist of dermoscopic images. Dermoscopy, while highly valuable for clinical practice, requires specialized equipment and trained dermatologists to capture detailed subsurface structures of the skin. Such resources are not always accessible in primary care or rural settings, limiting the applicability of dermoscopy-based models in real-world scenarios. In contrast, ordinary clinical or mobile camera images can be acquired easily and inexpensively without specialized tools, making them far more practical for widespread screening. In this study, we focus on diagnosing skin conditions from non-dermoscopic images captured with mobile devices, aiming to develop a system that can operate on simple outward appearances of skin lesions and thus improve accessibility to diagnostic support for preliminary screening.

%Obtaining such data is quite difficult. 
%as images captured outside clinical settings often vary widely in terms of resolution, quality, lighting conditions, and background noise. To address this gap,

In this study we curated a compiled dataset of non-dermoscopic skin images by collecting samples from diverse online sources. Images were gathered through extensive searches across publicly available platforms to capture a broad range of skin tones, lesion types, and photographic conditions. During this process, unusual, unclear, or low-quality images were carefully removed to ensure data reliability and consistency. Our curated dataset comprises more than 27,000 images spanning over 50 skin disease categories, making it one of the largest non-dermoscopic skin image collections available for research. This curated dataset constitutes one of the key contributions of our work, as it provides a valuable resource for developing and evaluating DL models on real-world, mobile-acquired skin images rather than specialized dermoscopic scans.

Developing an effective CAD tool requires models that can accurately capture both local lesion characteristics and global contextual patterns present in medical images. While CNNs have achieved remarkable success in skin lesion classification \cite{b7}, they primarily rely on local receptive fields, which often limit their ability to model long-range dependencies and holistic structural information. This becomes particularly important for complex lesions, where subtle patterns may span across different regions of the image (e.g., Pox and Psoriasis). Recently, transformer architectures have emerged as a powerful alternative in computer vision \cite{b8}. By leveraging a self-attention mechanism, transformers can capture both local and global relationships across image patches, making them well-suited for tasks that require context-aware feature extraction. Their introduction into medical imaging has demonstrated promising improvements in performance, positioning them as a strong candidate for advancing automated skin disease diagnosis. In this work, we investigate the use of DL models, particularly transformers, for the classification of non-dermoscopic skin images, aiming to provide a scalable solution for automated skin disease diagnosis.

In addition to that, interpretability is a critical requirement for CAD systems, particularly in medical applications where transparency and trust are essential. To address this, we employed the Gradient-weighted Class Activation Mapping (Grad-CAM) technique to visualize the regions of input images that most strongly influenced the model’s predictions. By generating heatmaps overlaid on the original skin images, Grad-CAM provides an intuitive understanding of whether the model is focusing on clinically relevant lesion areas rather than background artifacts. This not only enhances the reliability of the proposed framework but also offers valuable insights for clinicians and enables non-dermatologists to better assess and validate the diagnostic outcomes, thereby bridging the gap between DL models and practical medical use.

To summarize our contributions, we curated a large non-dermoscopic dataset of over 27,000 images across 50+ categories collected from diverse online sources, and the dataset is being actively updated to ensure broader coverage and diversity. We evaluated multiple DL models for skin lesion classification, with Transformer-based architectures (Swin) emerging as the best performers, reaching an accuracy of 81\%. By focusing on mobile-acquired images, our approach offers a practical solution for real-world scenarios where dermatologists and dermoscopy are not readily available. Finally, we incorporate Grad-CAM visualizations to enhance interpretability, enabling both clinicians and non-specialists to assess the diagnostic outcomes with greater confidence.

\section{Literature Review}

Skin diseases are among the most significant health concerns worldwide, yet conventional diagnostic approaches are often complex, costly, and inadequate for early detection. Consequently, automated classification of skin diseases using DL models has gained substantial importance. Early studies relied primarily on CNNs such as VGG16, ResNet, and Inception for lesion classification and segmentation \cite{b2,b7}. While CNNs have achieved promising results, they often struggle in complex scenarios, particularly when different conditions exhibit overlapping visual characteristics.

Recent studies has increasingly explored the use of Transformer-based models for medical image analysis. Vision Transformers (ViT) and their variants have shown superior performance in capturing global contextual features compared to CNNs \cite{b8}. However, these methods typically rely on large-scale datasets and are computationally demanding, making them challenging to apply in resource-constrained settings.

The availability of public datasets such as ISIC \cite{b4}, Derm7pt \cite{b6}, and HAM10000 \cite{b5} has fueled progress in this domain. While these collections provide valuable training resources, they are primarily dermoscopic, with a limited number of categories and rare conditions underrepresented. Moreover, they do not reflect real-world scenarios where only non-dermoscopic, mobile-acquired images are available. Such mobile datasets remain limited and scattered \cite{b10}, and to the best of our knowledge, no prior attempts have been made before this work to compile a large database of mobile-captured skin images.

Most existing work has focused on dermoscopic datasets with restricted disease categories, often limited to binary or small-scale multiclass classification. For example, Choudhary et al. \cite{b21} combined CNNs with GLCM-based features for binary classification on ISIC 2017, while Young et al. \cite{b22} used an Inception-based model for melanoma vs. nevus classification on ISIC 2018. Although effective, such binary approaches have limited clinical applicability. Arora et al. \cite{b23} compared 14 DL networks on ISIC 2018, with DenseNet201 achieving 82.5\% accuracy. Nakai et al. \cite{b25} incorporated self-attention into a bottleneck transformer using ISIC and HAM10000, improving feature extraction but still restricted to 6–7 classes. These studies highlight progress but remain limited in scope and clinical relevance.

Some researchers have attempted to merge multiple datasets to increase diversity. Rafay and Hussain proposed EfficientSkinDis, trained on 31 diseases from ISIC and Atlas Dermatology \cite{b26}, while Sadik et al. \cite{b27} combined HAM10000 with Dermnet. Hanum et al. \cite{b28} integrated five public datasets into 39 classes and evaluated attention-enhanced models. These studies combined dermoscopic datasets (e.g., ISIC) with mobile-acquired datasets (e.g., PAD-UFES-20, Dermnet) to increase dataset size.
However, merging data this way can introduce significant domain shift, as differences in resolution, lighting, and acquisition settings can cause models to learn dataset-specific artifacts instead of disease features. Without explicit domain adaptation, such approaches risk poor generalization. For practical deployment, evaluation must be performed on the intended acquisition modality.

%While this can improve diversity, it introduces a significant domain shift, since dermoscopic and clinical photographs differ in resolution, lighting, and acquisition setting. Without explicit domain adaptation, models trained on mixed datasets risk learning dataset-specific artifacts rather than disease-specific features. For real-world deployment, evaluation should therefore be performed on the intended acquisition modality.

To address these limitations, our study curates a dataset of over 50 skin diseases captured exclusively with mobile devices. This design enhances applicability for large-scale deployment, particularly in rural and resource-limited environments. Furthermore, we evaluate both CNN-based and Transformer-based models, and employ Grad-CAM to provide interpretability and insights into model decisions.

\section{Materials and Methods}

\subsection{Data Curation}

Obtaining non-dermoscopic skin images is quite challenging. Images captured outside clinical settings vary widely in resolution, quality, illumination, and background. We curated a dataset of more than \textbf{27,000 images} spanning over \textbf{50 skin disease categories}.
Images were collected from diverse online sources through extensive searches to capture a broad range of skin tones and lesion types. We also identified several public repositories that provide images for specific categories of skin conditions. For example, approximately 500 Monkeypox images were collected from one such repository \cite{b9}. The PAD-UFES-20 dataset contributed over 2,000 smartphone-acquired images across six skin lesion types \cite{b10}. In addition, the Dermatology Atlas served as a valuable source for obtaining images of many rare skin lesions \cite{b11}. However, not all retrieved images were of sufficient quality, and low-resolution or unclear samples were excluded during the curation process. This compilation provides one of the largest publicly reported collections of non-dermoscopic skin lesion images and forms the foundation for training and evaluating the DL models presented in this study. A summary of the curated dataset is provided in Table \ref{tab:dataset_summary}.

\begin{table*}[htbp]
\centering
\caption{Summary of the curated non-dermoscopic image dataset.}
\label{tab:dataset_summary}
\scriptsize
\begin{tabularx}{\textwidth}{|X|c|X|c|X|c|X|c|}
\hline
\textbf{Category} & \textbf{\#Images} & \textbf{Category} & \textbf{\#Images} & \textbf{Category} & \textbf{\#Images} & \textbf{Category} & \textbf{\#Images} \\ \hline
Actinic Keratosis & 1247 & Arsenic & 758 & Basal Cell Carcinoma & 1914 & Chickenpox & 482 \\ \hline
Cowpox & 330 & Eczema & 666 & xeroderma pigmentosum & 98 & Measles & 366 \\ \hline
Melanoma & 310 & Monkeypox & 1699 & Nevus & 244 & Normal & 1409 \\ \hline
Seborrheic Keratosis & 693 & Squamous Cell Carcinoma & 905 & acne & 342 & acne vulgaris & 394 \\ \hline
allergic contact dermatitis & 490 & cheilitis & 106 & dariers disease & 218 & dermatomyositis & 206 \\ \hline
drug eruption & 364 & ehlers danlos syndrome & 127 & erythema multiforme & 388 & fixed eruptions & 198 \\ \hline
folliculitis & 369 & fordyce spots & 118 & granuloma annulare & 236 & hailey hailey disease & 162 \\ \hline
herpes simplex & 292 & herpes zoster & 415 & hidradenitis suppurativa & 282 & hirsutism & 123 \\ \hline
ichthyosis vulgaris & 198 & impetigo & 277 & kaposi sarcoma & 121 & keratosis pilaris & 204 \\ \hline
leprosy & 291 & lichen planus & 330 & lichen sclerosus & 148 & lipoma & 224 \\ \hline
lupus erythematosus & 307 & molluscum contagiosum & 408 & mycosis fungoides & 172 & onychomycosis & 273 \\ \hline
parapsoriasis & 185 & pemphigus vulgaris & 196 & pityriasis rosea & 229 & pityriasis rubra pilaris & 156 \\ \hline
poikiloderma & 111 & porphyria cutanea tarda & 102 & psoriasis & 531 & purpura & 198 \\ \hline
rosacea & 246 & scabies & 405 & scleroderma & 167 & sebaceous cyst & 210 \\ \hline
stasis dermatitis & 192 & stevens johnson syndrome & 132 & tinea corporis & 288 & tinea cruris & 194 \\ \hline
tinea faciei & 208 & tinea manuum & 163 & tinea pedis & 260 & tinea unguium & 179 \\ \hline
toxic epidermal necrolysis & 106 & urticaria & 276 & verruca vulgaris & 301 & vitiligo & 412 \\ \hline
%Hand Foot Mouth Disease & 835 & Xanthomas & 119 & & & & \\ \hline
\end{tabularx}
\end{table*}

\subsection{Deep Learning Architectures}

In this study, we explored both CNNs and Transformer-based architectures for skin lesion classification. CNNs have been widely adopted in medical image analysis due to their strong ability to extract local features such as edges, textures, and color variations. However, their reliance on local receptive fields often limits their capacity to capture long-range dependencies and global contextual patterns. To address this, we also investigated Transformer-based architectures, which leverage self-attention mechanisms to model relationships across the entire image. Among these, ViT and Swin Transformer were considered as representative architectures, offering complementary strengths in capturing both local and global information. All models were trained and evaluated on our curated non-dermoscopic dataset to provide a comprehensive comparison. A list of all the DL models used in this study and their key features are summarized in Table \ref{tab:dl_models}.

\begin{comment}

\begin{table*}[t]
\centering
\caption{Deep learning architectures used in this study.}
\label{tab:dl_models}
\scriptsize
\begin{tabular}{|l|l|c|l|l|}
\hline
\textbf{Architecture} & \textbf{Type} & \textbf{Year} & \textbf{Key Feature} \\ \hline
VGG16 \cite{b12} & CNN-based & 2014 & Deep architecture with stacked $3 \times 3$ convolutions \\ \hline
InceptionV3 \cite{b14} & CNN-based & 2015 & Factorized convolutions and auxiliary classifiers \\ \hline
ResNet-50 \cite{b13} & CNN-based & 2016 & Residual connections for deep training \\ \hline
Xception \cite{b17} & CNN-based & 2017 & Depthwise separable convolutions \\ \hline
MobileNetV2 \cite{b16} & CNN-based & 2018 & Inverted residuals with linear bottlenecks \\ \hline
EfficientNet-B0 \cite{b15} & CNN-based & 2019 & Compound scaling of depth, width, and resolution \\ \hline
ViT \cite{b8} & Transformer-based & 2021 & Patch embeddings with self-attention \\ \hline
Swin Transformer \cite{b19} & Transformer-based & 2021 & Hierarchical vision transformer with shifted windows \\ \hline
ConvNeXt \cite{b18} & Transformer-inspired CNN & 2022 & CNN modernized with Transformer design principles \\ \hline
\end{tabular}
\end{table*}

\end{comment}

\begin{table}[b] % single-column table
\centering
\caption{Deep learning architectures used in this study.}
\label{tab:dl_models}
\scriptsize
\begin{tabular}{|p{1.9cm}|p{1.4cm}|p{0.5cm}|p{3.2cm}|}
\hline
\textbf{Architecture} & \textbf{Type} & \textbf{Year} & \textbf{Key Feature} \\ \hline
VGG16 & CNN & 2014 & Deep architecture with stacked $3 \times 3$ convolutions \\ \hline
InceptionV3 & CNN & 2015 & Factorized convolutions and auxiliary classifiers \\ \hline
ResNet-50 & CNN & 2016 & Residual connections for deep training \\ \hline
Xception & CNN & 2017 & Depthwise separable convolutions \\ \hline
MobileNetV2 & CNN & 2018 & Inverted residuals with linear bottlenecks \\ \hline
EfficientNet-B0 & CNN & 2019 & Compound scaling of depth, width, and resolution \\ \hline
ViT & Transformer & 2021 & Patch embeddings with self-attention \\ \hline
Swin Transformer & Transformer & 2021 & Hierarchical vision transformer with shifted windows \\ \hline
ConvNeXt & Transformer-inspired CNN & 2022 & CNN modernized with Transformer design principles \\ \hline
\end{tabular}
\end{table}

\subsection{Model Utilization}

For each model, we employed transfer learning with ImageNet-pretrained weights. These weights, obtained from training on the large-scale ImageNet dataset of over one million natural images, provide generic visual feature representations such as edges, textures, and shapes. By starting from pretrained weights rather than random initialization, the models benefit from strong transferable features, which is particularly valuable for medical imaging tasks where data per class is limited.

The convolutional and Transformer backbones were initially frozen, and a custom classification head was added, consisting of a global average pooling layer, one or two fully connected layers (128 or 256 neurons, depending on the architecture) with ReLU activation, dropout for regularization, and a final softmax. After training the head, we also experimented with unfreezing a few backbone layers and performing fine-tuning; however, this did not yield good improvements. The models started to overfit, and the performance dropped. This is likely due to the limited number of samples available per class, which made it difficult to optimize the large number of parameters in these deep architectures. Thus, training with frozen backbones and a custom head provided more stable and generalizable results.

\subsection{Interpretability with Grad-CAM}

Interpretability of model predictions is essential in medical applications, as it enables clinicians to better understand the decision-making process of DL models. To this end, we employed Grad-CAM to generate visual explanations of model predictions. Grad-CAM produces heatmaps by backpropagating the gradients of a target class to the final convolutional layers, highlighting the regions of the input image that most strongly influenced the prediction. These visualizations provide insight into whether the model is attending to clinically relevant areas of the skin lesion, rather than being biased by background or irrelevant artifacts. By presenting heatmaps alongside predictions, Grad-CAM enhances the transparency of the system and supports clinical decision-making, particularly in scenarios where dermatologists are not readily available.

\section{Experimental Evaluation}

\subsection{Experimental Setup}

In our curated dataset, some categories had too few images to provide meaningful training. They were excluded from the current experiments to ensure balanced and reliable evaluation. We plan to collect additional images for these underrepresented skin conditions and update the models accordingly in future work. For this study, the resulting dataset comprises 51 classes, representing a complex multiclass classification task. The dataset was split into 70\% training, 15\% validation, and 15\% testing. The validation set was used to monitor overfitting during training and selecting the best model checkpoint, while the final performance of the models was evaluated on the test set and is reported in this manuscript. Input images were resized according to the requirements of each model and normalized using ImageNet statistics. The overall training configuration is summarized in Table \ref{tab:training_setup}.

%Data augmentation, including random flips, rotations, and color jittering, was applied to improve generalization.

\begin{table}[b]
\centering
\caption{Training Configuration.}
\label{tab:training_setup}
\scriptsize
%\begin{tabular}{|l|l|}
\begin{tabularx}{\columnwidth}{|l|X|}
\hline
\textbf{Parameter} & \textbf{Value} \\ \hline
Dataset Split & 70\% Train, 15\% Validation, 15\% Test \\ \hline
Loss Function & Categorical Cross-Entropy \\ \hline
Image Sizes & 224×224 (most CNNs), 299×299 (InceptionV3, Xception), 384×384 (Transformers) \\ \hline
Optimizer & Adam \\ \hline
Learning Rate & $1 \times 10^{-4}$ \\ \hline
Batch Size & 32 \\ \hline
Epochs & Up to 50 (with early stopping) \\ \hline
Normalization & ImageNet statistics \\ \hline
Regularization & Dropout (0.3 or 0.4), Batch Normalization \\ \hline
Callbacks & EarlyStopping, ReduceLROnPlateau, ModelCheckpoint \\ \hline
%Augmentation & Random flips, rotations, color jittering \\ \hline
%Metrics & Accuracy, Precision, Recall, F1-score, AUC \\ \hline
Random Seed & Fixed for reproducibility \\ \hline
Hardware & Kaggle GPU (NVIDIA Tesla T4) \\ \hline
\end{tabularx}
\end{table}

\subsection{Performance Evaluation}

To comprehensively assess model performance, 5 different evaluation metrics suitable for multiclass classification were employed. Overall accuracy was reported as the primary performance indicator, while precision, recall, and F1-score were calculated to capture performance on individual classes. Since the dataset is imbalanced, we used weighted averaging, ensuring that class-level contributions were proportional to their sample sizes (denoted as Precision\textsubscript{w}). In addition to these conventional metrics, we also report the Matthews Correlation Coefficient (MCC), which is widely regarded as one of the most reliable metrics for imbalanced multiclass classification, as it considers all entries of the confusion matrix and provides a balanced evaluation even when class distributions are skewed. A confusion matrix was also generated to provide a detailed view of class-wise predictions and misclassifications.

\begin{comment}
    
\begin{equation}
Accuracy = \frac{\sum_{i=1}^{C} TP_i}{\sum_{i=1}^{C} (TP_i + FN_i)}
\end{equation}

\begin{equation}
Precision_i = \frac{TP_i}{TP_i + FP_i}
\end{equation}

\begin{equation}
Recall_i = \frac{TP_i}{TP_i + FN_i}
\end{equation}

\begin{equation}
F1_i = \frac{2 \times Precision_i \times Recall_i}{Precision_i + Recall_i}
\end{equation}

\begin{equation}
MCC = \frac{ \sum_{k} \sum_{l} \sum_{m} C_{kk} C_{lm} - C_{kl} C_{mk} }
{\sqrt{ \sum_{k} \left( \sum_{l} C_{kl} \right) \left( \sum_{k \neq l} C_{kl} \right) } 
\sqrt{ \sum_{k} \left( \sum_{l} C_{lk} \right) \left( \sum_{k \neq l} C_{lk} \right) }}
\end{equation}

\end{comment}

\subsection{Results}

The performance of all evaluated models on the test set is summarized in Table~\ref{tab:results_main}. Class-wise performance can be observed from the confusion matrices, which are available in the associated GitHub repository. Due to space limitations, they are not included in this manuscript.

Among the CNN-based architectures, ResNet-50, InceptionV3, and EfficientNet-B0 showed comparable performance, with ResNet-50 achieving the highest MCC (0.630). In contrast, MobileNetV2 performed noticeably worse, reflecting its lightweight design and limited representational capacity for this complex dataset. VGG16 also performed poorly. ConvNeXt, as a Transformer-inspired CNN, demonstrated moderate improvements over some CNN baselines but did not surpass the best-performing CNNs. Transformer-based models, however, achieved the strongest results overall, with the ViT and Swin Transformer significantly outperforming all CNN variants. In particular, the Swin Transformer obtained the highest accuracy, highlighting the benefit of self-attention mechanisms in capturing global contextual features for the challenging skin lesion classification task. 

The Swin Transformer comes in several variations (Tiny, Small, Base, and Large), which differ primarily in network depth \cite{b19}. Selecting the appropriate model size is crucial: smaller variants may lack sufficient capacity for complex datasets, while larger models risk overfitting when the number of class samples is limited. To balance these considerations, we evaluated two different variants. Among them, the Swin-Base model achieved the highest performance, with an F1 score of 81\% and an MCC score of 80\%.

These results also reflect the inherent complexity of the dataset, where many skin diseases share visually similar patterns, textures, and color distributions. Such inter-class similarity makes discrimination particularly challenging, even for advanced DL models. In addition, the limited number of samples per class further increases the risk of overfitting and hinders the models from learning subtle distinguishing features. Taken together, these factors highlight the difficulty of multiclass skin lesion classification and explain the limitation in performance across different architectures.

\begin{table}[htbp]
\centering
\caption{Test-set performance of different models.}
\label{tab:results_main}
\scriptsize
\begin{tabular}{|l|c|c|c|c|c|}
\hline
\textbf{Model} & \textbf{Accuracy} & \textbf{Precision\textsubscript{w}} & \textbf{Recall\textsubscript{w}} & \textbf{F1\textsubscript{w}} & \textbf{MCC} \\
\hline
VGG16 & 0.494 & 0.462 & 0.494 & 0.453 & 0.468 \\ \hline
InceptionV3 & 0.641 & 0.639 & 0.641 & 0.629 & 0.624 \\ \hline
ResNet-50 & 0.647 & 0.650 & 0.647 & 0.638 & 0.630 \\ \hline
Xception & 0.637 & 0.640 & 0.637 & 0.630 & 0.621 \\ \hline
MobileNetV2 & 0.427 & 0.387 & 0.427 & 0.386 & 0.397 \\ \hline
EfficientNet-B0 & 0.644 & 0.640 & 0.644 & 0.631 & 0.628 \\ \hline
ConvNeXt & 0.612 & 0.611 & 0.612 & 0.596 & 0.594 \\ \hline
ViT & 0.734 & 0.752 & 0.734 & 0.73 & 0.723 \\ \hline
Swin-Tiny & 0.775 & 0.791 & 0.775 & 0.78 & 0.766 \\ \hline
Swin-Base & 0.808 & 0.819 & 0.809 & 0.808 & 0.802 \\ \hline
\end{tabular}
\end{table}

\section{Discussion}

\subsection{Training Dynamics}

\begin{figure}[htbp]
\centering
\includegraphics[width=\linewidth]{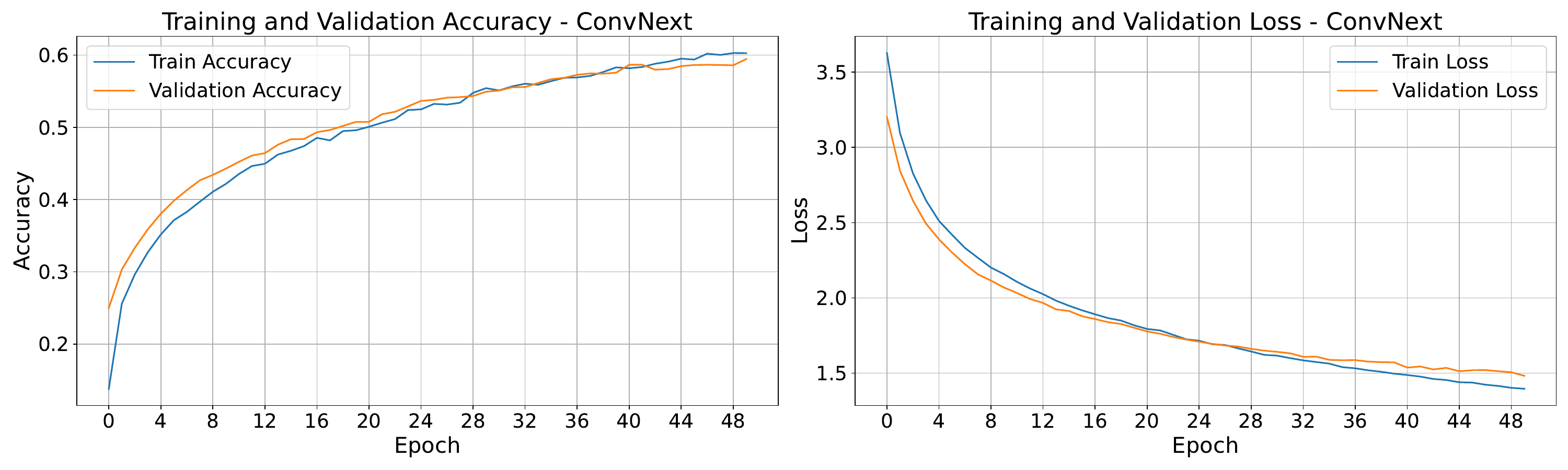}
\caption{Training and validation loss curves for ConvNeXt.} \vspace{0.5em}
\label{fig:convnext_loss}
%\end{figure}

%\begin{figure}[htbp]
  \centering
  \includegraphics[width=\linewidth]{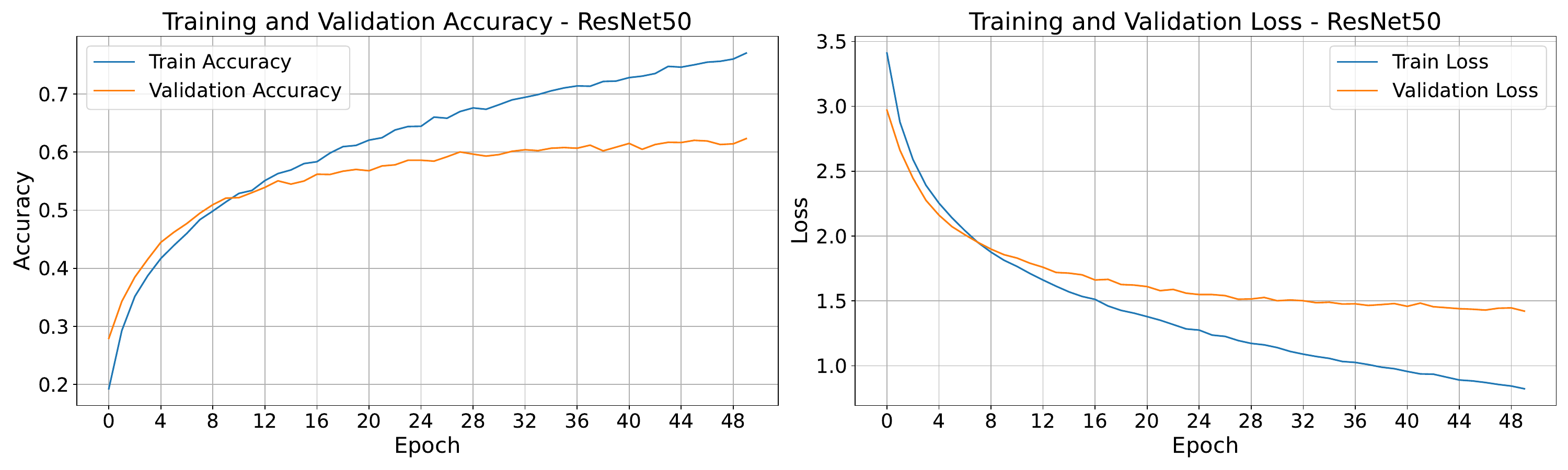}
  \caption{Training and validation loss curves for ResNet-50.} \vspace{0.5em}
  \label{fig:resnet50_loss}
%\end{figure}

%\begin{figure}[htbp]
\centering
\includegraphics[width=\linewidth]{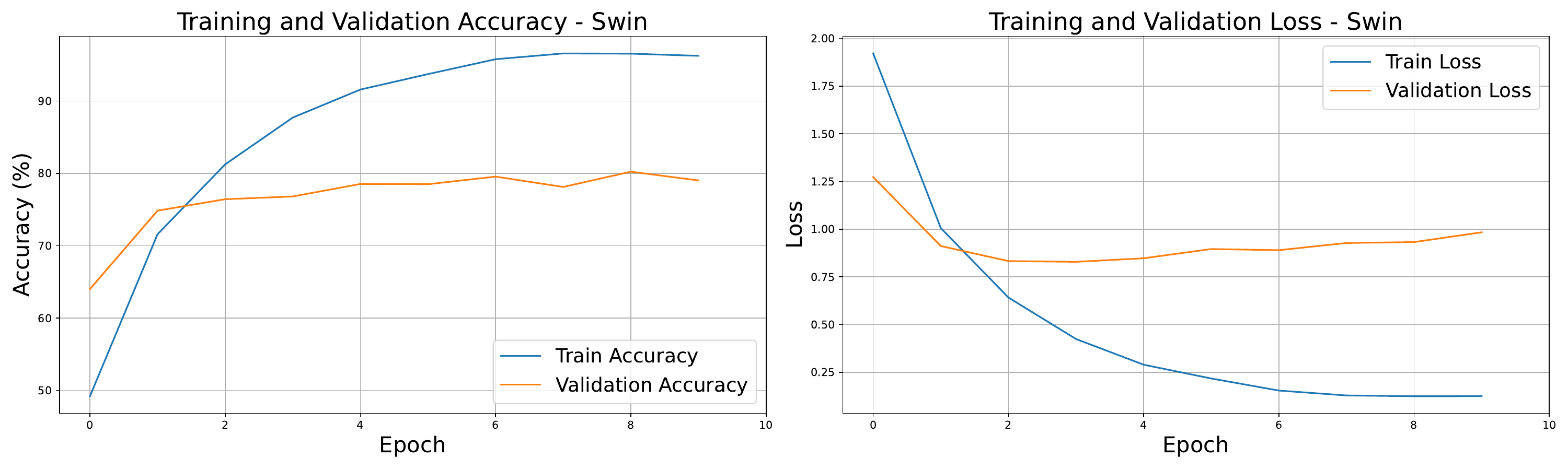}
\caption{Training and validation loss curves for Swin Transformer.}
\label{fig:swin_loss}

\end{figure}

To better understand the learning behavior of the models, we examined the training and validation loss curves of the architectures. Three representative cases are shown in this manuscript, while the complete set of curves is available in the GitHub repository. CNN-based models such as EfficientNet-B0 and ConvNeXt exhibited slow but stable convergence with minimal oscillations (Figure~\ref{fig:convnext_loss}); however, their performance plateaued at around 60\% accuracy. 
%This behavior can be attributed to their limited capacity to capture global context when trained on relatively small and highly diverse medical datasets. 
Other CNN variants showed signs of overfitting after approximately 20 epochs, although the validation loss continued to decrease gradually. This trend is illustrated in Figure~\ref{fig:resnet50_loss}, where the widening gap between training and validation curves reflects mild overfitting. In contrast, Transformer-based models converged rapidly within the first few epochs. For instance, the Swin Transformer reached nearly 95\% training accuracy within 10 epochs, while validation accuracy leveled off at around 80\% (Figure~\ref{fig:swin_loss}). Training was terminated early as validation loss began to increase, indicating overfitting. This fast convergence is a result of the self-attention mechanism, which enables the model to quickly capture long-range dependencies and global features. However, the limited dataset size prevented the Transformer from fully generalizing, leading to early overfitting and the need for early stopping.

\subsection{Misclassification Analysis}

To gain deeper insight into model limitations, we conducted a misclassification analysis using the confusion matrix obtained from different architectures. Since there are 51 categories, the heatmap is too large to fit in the manuscript. While the models usually achieved good overall accuracy on most classes, several categories were frequently confused by all of them due to their high visual similarity. For instance, basal cell carcinoma was often misclassified as squamous cell carcinoma and vice versa, both of which share overlapping color and texture characteristics. Similarly, eczema and psoriasis vulgaris were commonly intermixed, reflecting the subtle differences that even trained dermatologists may struggle to distinguish visually. In addition to disease similarity, limited sample availability for some categories also contributed to errors. Classes with fewer training examples, such as Kaposi sarcoma and lupus erythematosus, exhibited higher misclassification rates, often being predicted as more common conditions. 

\begin{figure}[b]
  \centering
  \subfloat[Correct predictions]{%
    \includegraphics[width=0.9\linewidth]{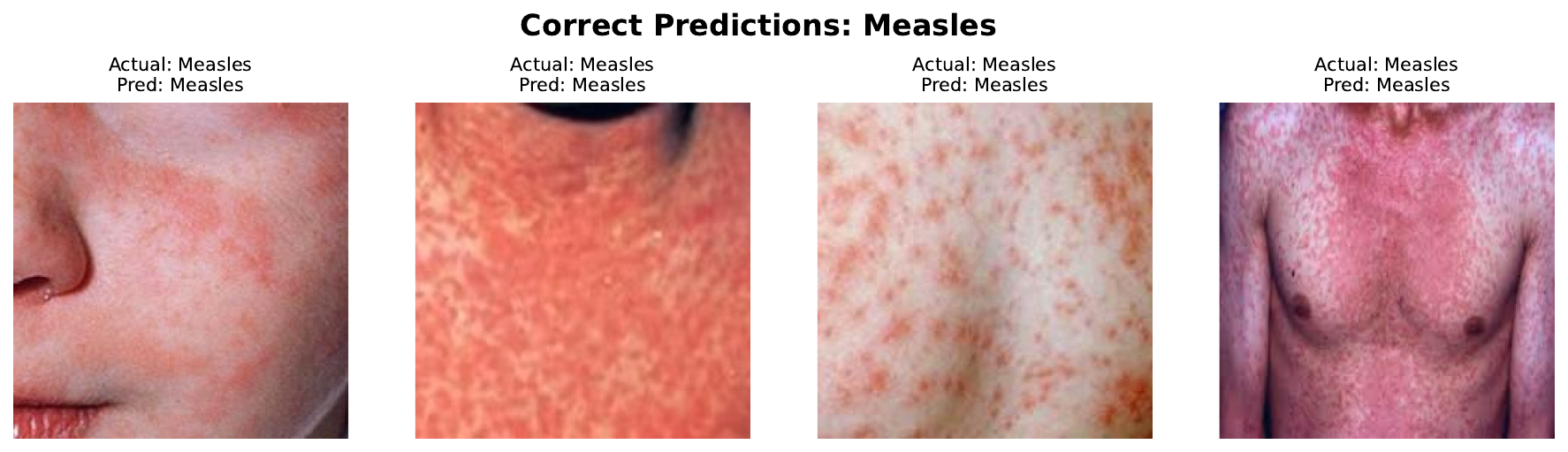}%
    \label{fig:swin_correct}
  }\vfill
  \subfloat[Misclassified cases]{%
    \includegraphics[width=0.9\linewidth]{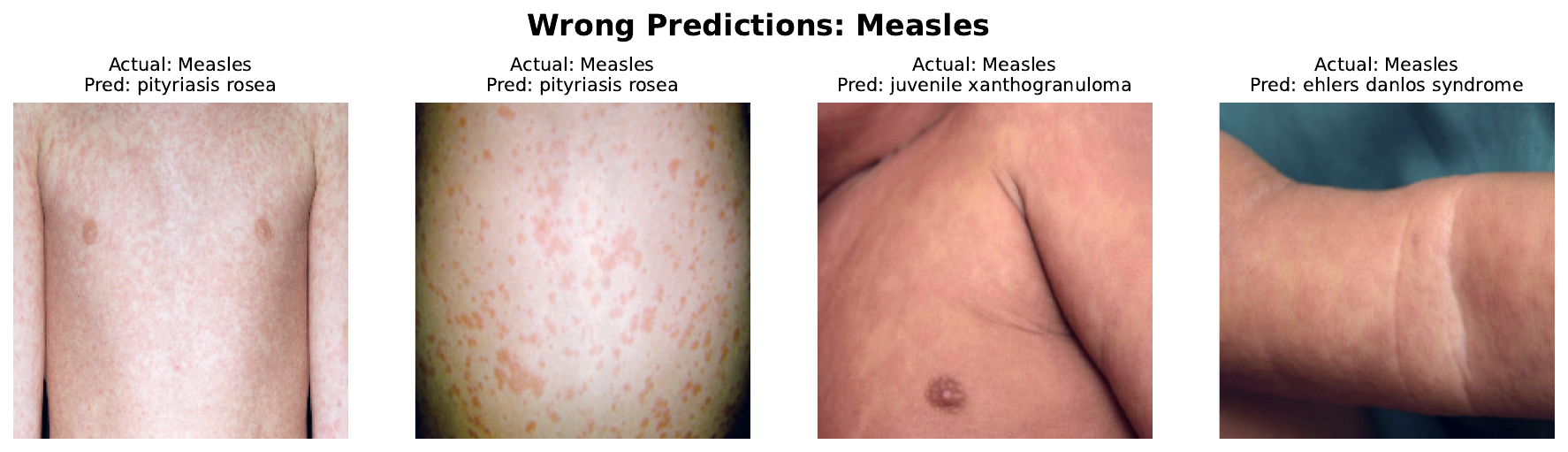}%
    \label{fig:swin_wrong}
  }
  \caption{Representative qualitative results of the Swin Transformer. (a) Correctly classified samples. (b) Misclassified cases.}
  \label{fig:swin_examples}
\end{figure}

Figures~\ref{fig:swin_correct} and \ref{fig:swin_wrong} show representative examples of correct and incorrect predictions made by the Swin Transformer for the `Measles' category. In the correctly classified cases, the model accurately identified measles across different patients (Figure~\ref{fig:swin_correct}). By contrast, the misclassified examples (Figure~\ref{fig:swin_wrong}) illustrate conditions where the model confused measles with visually similar skin diseases such as pityriasis rosea, juvenile xanthogranuloma, and Ehlers-Danlos syndrome. The number of training samples available for some of these categories is as low as ~100, which makes it difficult for the models to learn discriminative features and increases the likelihood of misclassification. These examples highlight both the strengths and limitations of the model, as some of the confusions involve conditions that are also challenging for dermatologists to differentiate solely based on clinical images.

\subsection{Fine Tuning and Data Augmentation}

\begin{figure}[b]
  \centering
  \includegraphics[width=\linewidth]{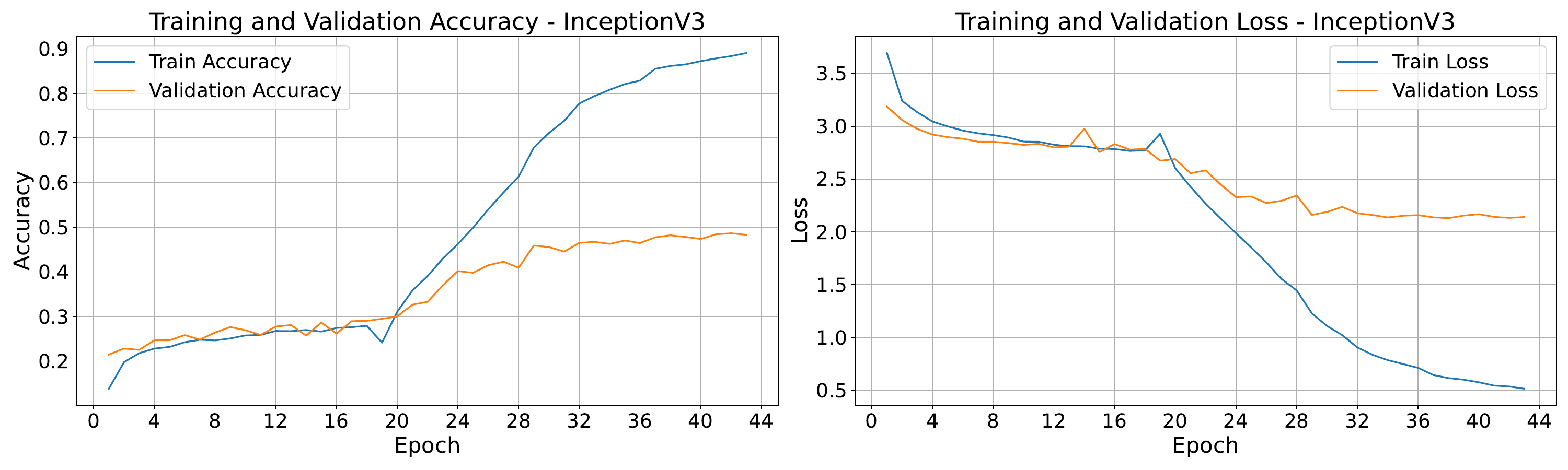}
  \caption{Training and validation accuracy/loss curves for InceptionV3 with fine-tuning.}
  \label{fig:inceptionv3_curves}
\end{figure}

We experimented with partial fine-tuning of the pretrained backbones. For the first 15 epochs, the backbone was kept frozen while only the classification head was trained, during which both training and validation accuracy improved steadily. After this phase, the last few layers of the backbone were unfrozen, and training continued. As shown in Figure~\ref{fig:inceptionv3_curves}, once fine-tuning began, the model quickly started to overfit: training accuracy increased further, but validation accuracy plateaued and validation loss diverged. This indicates that, under data-constrained conditions, fine-tuning additional layers does not yield further improvements and may actually harm generalization. The likely reason is that the limited number of samples per class (some categories having barely ~100 images) is insufficient to effectively update the large number of parameters in these networks, leading to overfitting and disruption of the useful generic features obtained from ImageNet pretraining. Similar observations have been reported in other medical imaging studies, where transfer learning with frozen backbones and a lightweight classifier head has proven more effective than deeper fine-tuning \cite{b20}.

We also experimented with both conventional data augmentation techniques and GAN-based synthetic image generation. Simple data augmentation did not improve performance. The GAN-generated images lacked sufficient quality, likely due to the limited number of samples available per class, which hindered the model’s ability to generalize effectively. These findings suggest that generative augmentation may require larger and more balanced datasets to produce meaningful gains in this context.

\subsection{Interpretability with Grad-CAM}

\begin{figure}[htbp]
  \centering
  \subfloat[Milia]{%
    \includegraphics[width=0.95\linewidth]{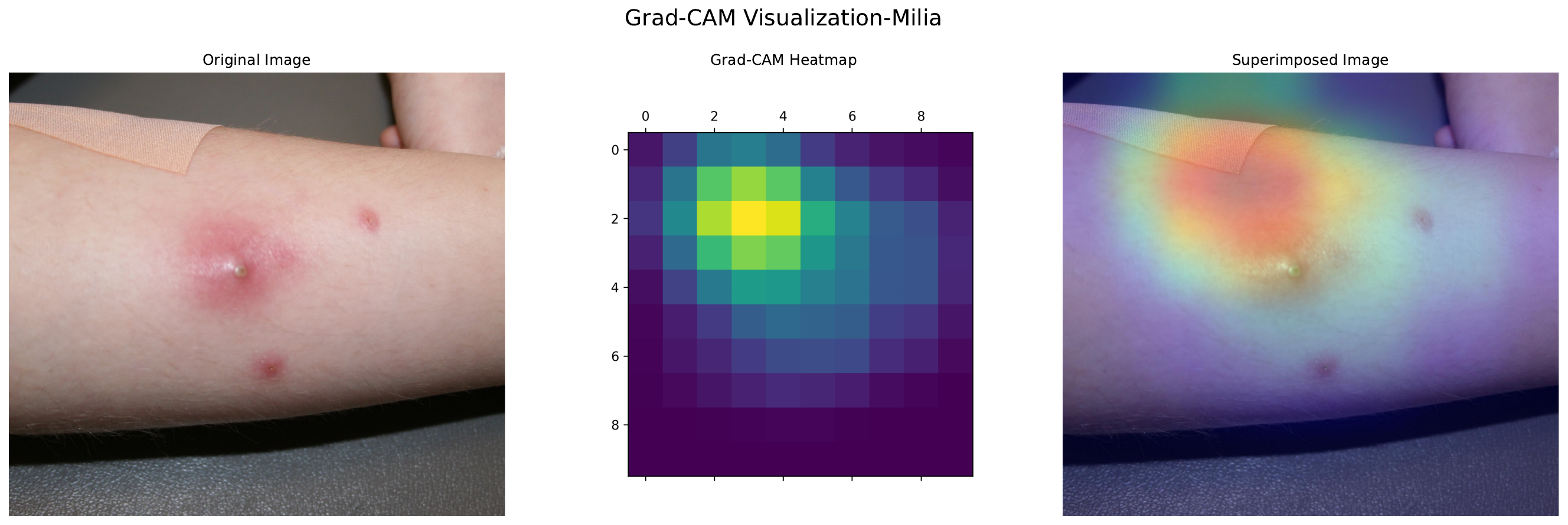}%
    \label{fig:gradcam_milia}
  }\vfill
  \subfloat[Keloid]{%
    \includegraphics[width=0.95\linewidth]{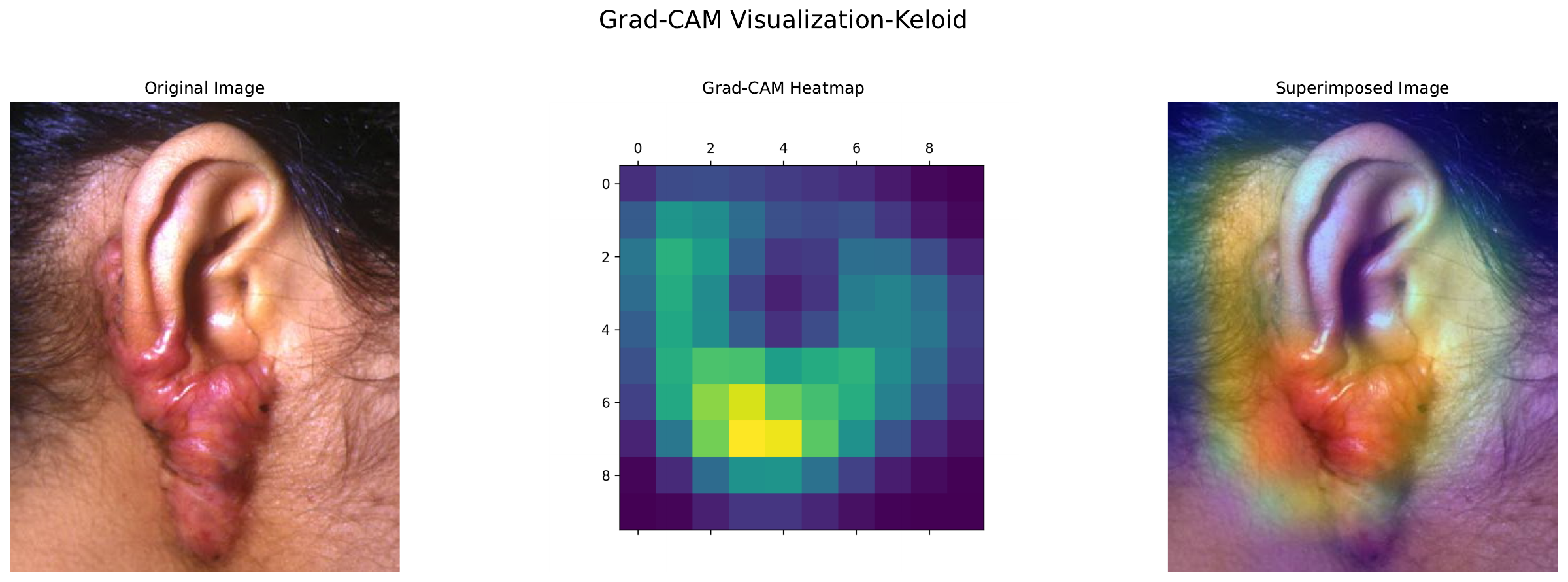}%
    \label{fig:gradcam_keloid}
  }
  \caption{Grad-CAM visualizations for two representative cases. 
}
  \label{fig:gradcam_examples}
\end{figure}

To enhance interpretability, we applied Grad-CAM to visualize the regions of input images that most influenced the model’s predictions. Figure~\ref{fig:gradcam_examples} shows representative Grad-CAM heatmaps generated for some samples. The highlighted regions closely align with the visible lesions, indicating that the model is attending to clinically relevant areas rather than unrelated background features. These visualizations not only provide transparency into the decision-making process but also support clinical usability by helping both dermatologists and non-specialists assess whether the model is basing predictions on meaningful image features.

%For misclassified cases, however, the heatmaps reveal scattered or misplaced attention, suggesting that the model was unable to focus consistently on the lesion of interest. 

\subsection{Limitation}

Although the proposed framework demonstrates promising performance in skin lesion classification, several limitations remain. First, the dataset, while relatively large, still contains only around 100 images for some categories, which restricts the ability of models to learn discriminative features and leads to higher misclassification rates for underrepresented conditions. Second, all images were collected from online sources, which may introduce variability in image quality, resolution, and acquisition conditions. Such heterogeneity could affect the generalizability of the trained models when applied to standardized clinical settings. Third, the study relied exclusively on image-based features; in real-world practice, dermatologists often consider patient history, dermoscopic patterns, and histopathological examinations, which were not available in this work.

%In future work, we plan to expand the dataset by incorporating more samples for rare categories and collecting images from controlled clinical environments. Integrating multimodal patient data could further improve diagnostic accuracy and robustness. Additionally, exploring different augmentation techniques may help mitigate overfitting in categories with limited samples. Finally, extending interpretability beyond Grad-CAM, such as through attention rollout or SHAP-based methods, could provide deeper insights into model decision-making and increase trust.

\section{Conclusion}

In this study, we presented a comprehensive framework for automated skin disease classification using a large, curated dataset of mobile-acquired images covering over 50 categories. Unlike most prior works that rely on dermoscopic images and limited disease classes, our dataset reflects real-world conditions and enhances the applicability of deep learning solutions in rural and resource-constrained settings. We evaluated multiple CNN-based and Transformer-based architectures, demonstrating that Transformer models, particularly Swin Transformer, achieved superior performance by leveraging their ability to model long-range dependencies and complex feature interactions. To ensure transparency and trust, we incorporated Grad-CAM visualizations, which provided interpretability of the predictions.

Despite these promising results, challenges remain due to inter-class similarities, data imbalance, and limited samples for rare categories. These factors contribute to misclassifications and hinder generalization. Future work will focus on expanding the dataset with more diverse and balanced samples, incorporating multimodal clinical data, and exploring advanced domain adaptation techniques to bridge variations across acquisition modalities. With these improvements, AI-assisted skin disease diagnosis can move closer to practical deployment, supporting dermatologists and non-specialists alike in achieving earlier and more reliable detection of skin conditions.

\section*{Acknowledgment}

The curated dataset, source code, and associated files are publicly available at: https://github.com/newaz-aa/skin\_lesion\_classification\_DL

%\section*{References}

\end{document}